\pdfoutput=1

\documentclass[11pt]{article}

\usepackage[final]{acl}

\usepackage{times}
\usepackage{latexsym}

\usepackage[T2A,T1]{fontenc}

\usepackage[utf8]{inputenc}
\usepackage[russian,english]{babel}

\usepackage{microtype}

\usepackage{inconsolata}

\usepackage{graphicx}

\usepackage{tcolorbox}
\usepackage{booktabs, multirow} 
\usepackage{soul}
\usepackage{pgfplots}
\usepackage{filecontents}
\usepgfplotslibrary{groupplots}
\usepackage{helvet}
\usepackage[eulergreek]{sansmath}
\pgfplotsset{
  compat=newest,
}
\usepackage{pgfplotstable}
\usepgfplotslibrary{colorbrewer}

\usepackage{todonotes}

\definecolor{tartunlp_violet}{HTML}{7268D8}
\definecolor{tartunlp_red}{HTML}{EF6650}
\definecolor{tartunlp_yellow}{HTML}{E0B12B}
\definecolor{tartunlp_black}{HTML}{282828}
\definecolor{tartunlp_green}{HTML}{4DB6AC}
\definecolor{tartunlp_blue}{HTML}{3185FF}
\definecolor{tartunlp_gray}{HTML}{9B9B9B}

\title{LLMs for Extremely Low-Resource Finno-Ugric Languages}

\author{Taido Purason\textsuperscript{*}, \, Hele-Andra Kuulmets\textsuperscript{*}, \, Mark Fishel\\
  Institute of Computer Science \\
  University of Tartu, Estonia \\
  \texttt{\{taido.purason, hele-andra.kuulmets, mark.fisel\}@ut.ee} \\
  }

\begin{document}
\maketitle
\begin{abstract}
The advancement of large language models (LLMs) has predominantly focused on high-resource languages, leaving low-resource languages, such as those in the Finno-Ugric family, significantly underrepresented. This paper addresses this gap by focusing on Võro, Livonian, and Komi. We cover almost the entire cycle of LLM creation, from data collection to instruction tuning and evaluation. Our contributions include developing multilingual base and instruction-tuned models; creating evaluation benchmarks, including the \textsc{smugri-MT-bench} multi-turn conversational benchmark; and conducting human evaluation. We intend for this work to promote linguistic diversity, ensuring that lesser-resourced languages can benefit from advancements in NLP.
\end{abstract}
\renewcommand{\thefootnote}{\fnsymbol{footnote}}
\footnotetext[1]{Equal contribution}
\def\thefootnote{\arabic{footnote}}
\section{Introduction}

Large language models (LLMs) have recently demonstrated unprecedented flexibility in responding to unstructured text queries \citep[etc]{openai2024gpt4,touvron2023llama}. However, their development requires high amounts of training material: while the (continued) pre-training stage only needs raw text, instruction tuning relies on sets of instructions, which are much more expensive to obtain. The challenge is exacerbated for endangered languages, where both the availability of data as well as the number of speakers are severely limited.

We present a case study in developing LLMs for extremely low-resource (XLR) languages, focusing on three Finno-Ugric (\textsc{smugri}\footnote{\emph{Finno-Ugric} translates to Estonian as \textit{\textbf{s}oo\textbf{m}e-\textbf{ugri}}, to Finnish as \textit{\textbf{s}uo\textbf{m}alais-\textbf{ugri}laiset}, to Võro as \textit{\textbf{s}oo\textbf{m}õ-\textbf{ugri}}, and to Livonian as \textit{\textbf{s}ūo\textbf{m}õ-\textbf{ugr}õ}, which is why we refer to it as \textsc{smugri}.}) languages: Livonian, Võro and Komi. According to \citet{joshi-etal-2020-state}, Livonian is in the lowest category out of 6 (0 / \textit{The Left-Behinds}) while Võro and Komi are in the second-lowest (1 / \textit{The Scraping-Bys}). Developing LLMs and other tools for these languages is thus a significant challenge, as well as a vital step for ensuring their digital survival and support.

\begin{table}[!tp]\centering
\small
\addtolength{\tabcolsep}{-1pt}
\begin{tabular}{l@{}rcccc}\toprule
 &\textbf{Script} &\textbf{Code} &\textbf{Class} &\textbf{Speakers} & \textbf{Status} \\\midrule
\textbf{Livonian}  &Latin &liv &\textbf{0}/5&\char`\~30\textsuperscript{\textdagger}  & \textcolor{tartunlp_red}{CE} \\
\textbf{Võro} &Latin &vro  &\textbf{1}/5  &\char`\~100K& \textcolor{tartunlp_yellow}{DE} \\
\textbf{Komi}  &Cyrillic &kpv &\textbf{1}/5 &\char`\~160K  & \textcolor{tartunlp_yellow}{DE}\\
\bottomrule
\end{tabular}
\caption{Language statistics of the targeted languages. The \textit{class} column indicates the amount of data available (on a scale from 0 to 5) as defined by \citet{joshi-etal-2020-state}. \textit{Status} according to \citet{moseley2010atlas}: \textcolor{tartunlp_yellow}{DE} - \textit{\textcolor{tartunlp_yellow}{definitely endangered}}; \textcolor{tartunlp_red}{CE} - \textit{\textcolor{tartunlp_red}{critically endangered}}. \textdagger - people able to communicate in Livonian\protect\footnotemark. }\label{tab:lang-stats}
\end{table}
\footnotetext{\href{http://www.livones.net/en/valoda/the-livonian-language}{www.livones.net/en/valoda/the-livonian-language}}

Our contributions cover the full cycle of LLM development, including continued pre-training and instruction tuning, as well as benchmark creation, and both automatic and manual evaluation. During pre-training and instruction-tuning, we make use of cross-lingual transfer from related higher-resourced languages and parallel translation data. Additionally, we rely on the small amounts of available parallel data for the included languages in order to develop intermediate translation functionality, which is then applied to machine-translate instructions into the target languages.

We also describe our significant manual effort. First, this includes extending two existing benchmarks to these target languages (multiple-choice question-answering and topic classification), one of which required additional manual translation. Second, it involves creating a new parallel multi-turn benchmark for these XLR target languages. The extended benchmarks enable automatic evaluation of language models on the target languages, while the multi-turn benchmark allows us to assess the real-life usefulness of instruction-tuned models.

We use the multi-turn benchmark to conduct extensive human evaluation, comparing our models to GPT-3.5-turbo. Translation, benchmark creation, and human evaluation were all carried out by native speakers of Komi and Võro. As there are no native speakers of Livonian, this work was handled by fluent speakers.

Evaluation on multiple-choice QA benchmarks indicates that our instruction-tuned models either outperform or are on par with strong proprietary baselines (GPT-3.5-turbo and GPT-4-turbo), for Livonian and Komi. Extensive human evaluation on the multi-turn benchmark further supports these findings. However, both automatic and human evaluations reveal that our models slightly underperform on Võro compared to proprietary models, likely due to Võro's close similarity to Estonian, a language in which proprietary models excel. Nonetheless, human evaluation shows that our models significantly outperform proprietary ones in terms of naturalness for Võro, Komi, and Livonian.

We publish the training implementation, evaluation benchmarks, and models\footnote{\href{https://github.com/TartuNLP/smugri-llm}{https://github.com/TartuNLP/smugri-llm}}.

\section{Background and Related Work}
\subsection{Finno-Ugric Languages} \label{sec:smugri-background}

Finno-Ugric languages belong to the Uralic language family and are spoken primarily in regions surrounding the Baltic Sea and the Ural Mountains. Most of these morphologically complex languages are considered low-resource or extremely low-resource (XLR), with Finnish, Hungarian, and Estonian being the most well-resourced. 
In this work, we focus on three XLR Finno-Ugric languages: Võro, Komi, and Livonian. These languages differ in both script and resource availability (see Table~\ref{tab:lang-stats}), which includes not only textual data but also the number of speakers. For example, Livonian has only 30 speakers, yet its community is highly active in preserving and revitalizing the language, exemplified by the establishment of the UL Livonian Institute in 2018.

The communities of Finno-Ugric language speakers have actively contributed to the development of modern NLP tools for their languages, including core NLP technologies, such as foundation models \citep{tanvir2020estbert, kuulmets2024teaching, luukkonen-etal-2023-fingpt, luukkonen2024poro}, which underpin many advanced NLP applications. Additionally, practical tools like machine translation systems  \citep{yankovskaya-etal-2023-machine,tars-etal-2022-teaching,tars2021extremely, DBLP:conf/acl/RiktersTTEF22} and speech synthesis technologies \citep{ratsep-fishel-2023-neural} have been developed to further support the use of these languages.

\textbf{Supporting languages.} To address the extreme data scarcity during continued pre-training, we include additional languages into the training data. First, we include Estonian and Finnish, which belong to the same Balto-Finnic subgroup as Võro and Livonian. Second, Latvian, due to its significant influence on Livonian and the fact that many Livonian speakers also speak Latvian. Third, Russian, because of its strong influence on Komi and the proficiency of many Komi speakers in Russian.

\subsection{Multilingual LLMs}

Multilingual LLMs are widely explored to expand the language coverage of LLMs. Traditional approaches involve training models from scratch \citep{luukkonen2024poro, luukkonen-etal-2023-fingpt, wei2023polylm, kudugunta2023madlad}. However, adapting pre-trained English-centric models to other languages through continued pre-training has also yielded promising results across various languages \cite{csaki2024sambalingo, dou2024sailor, rijgersberg2023geitje, lin2024mala500, andersland2024amharic, basile2023llamantino, owen2024komodo, cui2024efficient, cui2024rethinking, zhao2024llama, etxaniz-etal-2024-latxa}. In the context of Finno-Ugric languages, the most relevant works to ours include \citet{kuulmets2024teaching}, who adapted LLaMA-2 7B for Estonian, and \citet{luukkonen-etal-2023-fingpt}, who adapted BLOOM \citep{workshop2023bloom176bparameteropenaccessmultilingual} for Finnish. Additionally, \citet{luukkonen-etal-2023-fingpt} demonstrate that continued pre-training of BLOOM outperforms Finnish models trained from scratch, emphasizing the advantages of this approach.

The development of multilingual LLMs often employs techniques that enhance model quality. Common practices include incorporating parallel data into the pre-training phase \citep{luukkonen2024poro, owen2024komodo, wei2023polylm} and utilizing curriculum learning \citep{wei2023polylm}.

\subsection{Instruction Tuning}
Previous works have also explored a variety of cross-lingual techniques for teaching the models to follow instructions \cite{li2023align, zhu2023extrapolating, zhang2024enhancing, chai2024xcot, ranaldi-pucci-2023-english, chen2023breaking}. \citet{zhang2024enhancing} creates model answers to instructions in a high-resource/high-quality language, which are then translated and code-switched. Adding translation data during instruction-tuning has also been widely explored \cite{cui2024efficient, kuulmets2024teaching, zhu2023extrapolating, zhang2024enhancing, ranaldi-pucci-2023-english, chen2023breaking}. \citet{kuulmets2024teaching} find that using a diverse set of instructions in English can increase performance in Estonian tasks.

\subsection{Evaluation}

Common approaches to evaluating the multilingual capabilities of LLMs include using existing cross-lingual benchmarks \cite{ahuja-etal-2023-mega, ahuja2023megaverse} or translating English benchmarks into target languages through either machine translation \citep{lai-etal-2023-okapi} or manual translation \citep{shi2022language}. However, extending the evaluation of conversational capabilities to other languages is more complex, as a gold standard requires the involvement of human annotators \citep{touvron2023llama}. Human annotators are essential for both the recently popularized method of ranking models using the Elo rating system \citep{zheng2024judging} and the traditional method of pairwise comparison of answers from different models to predefined prompts \citep{zheng2024judging, touvron2023llama}.

An alternative line of research explores using LLMs as potential replacements for human annotators \cite{zheng2024judging, kim2023prometheus, kim2024prometheus}. While strong LLMs can effectively serve as substitutes for human annotators in English, their capabilities in non-English languages remain unclear. \citet{hada-etal-2024-large} investigate this across eight high-resource non-English languages, finding a bias in GPT-4-based evaluators toward assigning higher scores. To our knowledge, the behavior of LLM judges on XLR languages, including Finno-Ugric languages, has not been systematically studied.

\section{Experimental Setup}

\begin{table}[!htp]\centering
\scriptsize
\begin{tabular}{lrrrrrr}\toprule
\multirow{2}{*}{\textbf{Lang}} &\multirow{2}{*}{\textbf{Characters}} &\multicolumn{4}{c}{\textbf{Sampled Characters}} \\\cmidrule{3-6}
&  &\textbf{Stage 1} &\textbf{Stage 2} &\textbf{Total} &\textbf{Ratio} \\\midrule
\textbf{LIV} &2.6M  &- &10.3M &10.3M &4.00 \\
\textbf{VRO} &14.0M  &- &56.1M &56.1M &4.00 \\
\textbf{KPV} &578.9M  &- &1.4B &1.4B &2.48 \\
LV &27.8B  &3.0B &300.0M &3.3B &0.12 \\
ET &32.6B  &8.2B &300.0M &8.5B &0.26 \\
FI &114.0B  &7.6B &300.0M &7.9B &0.07 \\
RU &>1T  &2.7B &300.0M &3.0B &<0.01 \\
EN &>1T  &2.7B &300.0M &3.0B &<0.01 \\
\bottomrule
\end{tabular}
\caption{Training dataset composition. All of the data for Livonian is sentence-level, for other languages, the data is document-level.}\label{tab:train-ds}
\end{table}

\subsection{Continued Pre-training}\label{sec:training-data}

We take the approach of adapting the English-centric Llama-2 7B model \citep{touvron2023llama} to the target languages through full fine-tuning. Given our computational budget limitations, we employ a two-stage training strategy. In the first phase, we continue pre-training Llama-2 7B on higher-resource languages Finnish, Estonian, Russian and Latvian. In the second phase, we focus on teaching the model the XLR target languages resulting in \textbf{Llama-SMUGRI}. The training hyperparameters are detailed in Appendix~\ref{sec:training-params}.

\textbf{Stage 1: Learning supporting languages.} In the first step, we continue pre-training Llama-2 7B \cite{touvron2023llama} on higher-resource languages Estonian, Finnish, English, Latvian, and Russian. We allocate a training budget of 10 billion tokens and sample documents from CulturaX \cite{nguyen2023culturax}, with 32\%, 32\%, 12\%, 12\%, and 12\% probability of choosing the document from the respective language.

\textbf{Stage 2: Learning Võro, Komi and Livonian.} The second stage of continued pre-training focuses on enhancing the understanding and generative capabilities for XLR languages. We employ a character-based budget to ensure a balanced representation of languages in the training dataset. This budget is set at 3 billion characters, with 50\% allocated to sampling Võro, Komi, and Livonian using Unimax with N=4 \cite{chung2023unimax}, and the remaining 50\% uniformly distributed among the supporting languages to maintain the quality achieved in Stage 1. The N was chosen based on perplexity from the held-out validation set (see Appendix~\ref{sec:unimax}). The Komi documents are sourced from FU-LAB's Komi corpus\footnote{\href{http://wiki.fu-lab.ru/index.php/\%D0\%AD\%D0\%BB\%D0\%B5\%D0\%BA\%D1\%82\%D1\%80\%D0\%BE\%D0\%BD\%D0\%BD\%D0\%B0\%D1\%8F_\%D0\%B1\%D0\%B0\%D0\%B7\%D0\%B0_\%D0\%BA\%D0\%BE\%D0\%BC\%D0\%B8_\%D1\%82\%D0\%B5\%D0\%BA\%D1\%81\%D1\%82\%D0\%BE\%D0\%B2}{\foreignlanguage{russian}{http://wiki.fu-lab.ru/index.php/Электронная\_\\база\_коми\_текстов}}}. The Livonian dataset consists of sentence-level data from \citet{DBLP:conf/acl/RiktersTTEF22}, while Võro dataset is compiled from various pre-existing corpora as well as data we have scraped. A more detailed overview of Võro dataset can be found in Appendix~\ref{sec:vro-composition}.

\textbf{Stage 2 + parallel: making use of parallel translation data.} To investigate the role of parallel translation examples in the pre-training data, we incorporate translation examples formatted into various templates, accounting for up to 1\% of the Stage 2 character budget (159,712 sentence pairs). We use Unimax with N=1 to balance the budget between language pairs. For further details on the parallel data, we refer the reader to Appendix~\ref{sec:parallel-data}. This stage yields the final base model we refer to as \textbf{Llama-SMUGRI}.

\subsection{Instruction-tuning}
\textbf{Supporting instructions.} We utilize existing instruction-tuning datasets across multiple languages. For English, Russian, and Finnish, we use Aya \cite{singh2024aya}, and the highest-rated conversation paths of OASST-2 \cite{kopf2023openassistant}. Additionally, we sample 20,000 Estonian instructions from Alpaca-est \cite{kuulmets2024teaching}. Following \citet{kuulmets2024teaching}, we include 5,000 instructions from the FLAN-V2 \cite{pmlr-v202-longpre23a} TULU mixture \cite{wang2023how} and 20,000 examples from Alpaca-GPT-4 \cite{peng2023instruction}, to improve cross-lingual knowledge transfer from high-quality English instructions. We refer to this instruction mixture as \texttt{SupInst} (Supporting Instructions). Further details are listed in Appendix~\ref{sec:instruction-tuning-details}.

\textbf{XLR Language Instructions.} Due to LLMs' insufficient capabilities in XLR languages, it is not feasible to create Alpaca-style instructions directly. Consequently, we create instruction datasets for Võro, Livonian and Komi by translating 1000 examples per language from Alpaca-style instruction datasets into these languages. An external system, Neurotõlge\footnote{\href{https://neurotolge.ee/}{https://neurotolge.ee/}} \citep{yankovskaya-etal-2023-machine}, is used for translation. While Võro and Livonian are translated directly from Alpaca-est \cite{kuulmets2024teaching}, Komi is generated by first translating Alpaca-GPT-4 \cite{peng2023instruction} into Russian using \texttt{GPT-3.5-turbo}, and then translating that result into Komi with Neurotõlge. We refer to this dataset as \texttt{TrAlpaca}. 

To investigate a scenario where a translation model is unavailable, we explore handling translating Alpaca instructions to XLR languages by fine-tuning our base model (Llama-SMUGRI) for the translation task (discussed in \S\ref{sec:translation-tuning}) resulting in \texttt{LLMTrAlpaca} instructions. This is similar to self-translate-train \cite{ri2024selftranslatetrainenhancingcrosslingualtransfer} and self-translate \cite{etxaniz-etal-2024-multilingual} concepts with the difference that we add a fine-tuning step to obtain an LLM specialized for translation. Further instruction-translation details and human evaluation of translations are discussed in Appendix~\ref{sec:instruction-translation}.

\textbf{Translation instructions.} We augment the general instructions with translation task instructions for Võro, Livonian, and Komi, using 250 examples per direction. We refer to these translation task instructions as \texttt{TrInst} (see Appendix~\ref{sec:parallel-data} for dataset overview).

\subsection{Translation-tuning}\label{sec:translation-tuning}

Adapting general-purpose LLMs for the machine translation task has been shown to yield competitive results compared to dedicated MT systems \cite{xu2023paradigm, kuulmets2024teaching}. Therefore, we fine-tune our base model on available translation data by sampling up to 100,000 sentence pairs from each language pair (see Appendix~\ref{sec:parallel-data} for further details) to compare the quality of our model to using an MT system. We call this configuration \texttt{TrTuning}.

\section{Benchmarks}
\setlength{\tabcolsep}{3pt}
\begin{table}[!htp]\centering
\scriptsize
\begin{tabular}{@{}llrl@{}}\toprule
\textbf{Benchmark}& &\textbf{Size} &\textbf{Type} \\\midrule
 MT-bench-SMUGRI  & \textbf{[new]} &80 &multi-turn questions \\
 Belebele-SMUGRI  & \textbf{[extended]} &127 &multi-choice QA \\
 SIB-SMUGRI  & \textbf{[extended]} &125 &topic classification \\
FLORES-SMUGRI  & \citet{yankovskaya-etal-2023-machine} & 250 &translation \\
\bottomrule
\end{tabular}
\caption{Test benchmarks for Komi, Võro, and Livonian. 
}\label{tab:collected-benchmarks}
\end{table}
\setlength{\tabcolsep}{6pt}

\subsection{Automatic Evaluation}

\textbf{Existing benchmarks.} From the existing benchmarks we use FLORES-SMUGRI \citep{yankovskaya-etal-2023-machine} machine translation benchmark. It includes the first 250 sentences of FLORES-200 \cite{nllbteam2022language, goyal-etal-2022-flores} translated into several Finno-Ugric languages. 

\textbf{New benchmarks.} We extend the topic classification benchmark SIB-200 \cite{adelani-etal-2024-sib} and the multiple-choice QA benchmark Belebele \cite{bandarkar2023belebele} to include Livonian, Võro, and Komi. Both SIB-200 and Belebele build on top of FLORES-200 and, therefore, can be extended to Livonian, Võro, and Komi using translations by \citet{yankovskaya-etal-2023-machine}. We align these translations with sentences in SIB-200 and with paragraphs in Belebele. We then manually translate the questions and answer choices in Belebele into the target languages, as FLORES-200 does not contain these components.  Table~\ref{tab:collected-benchmarks} shows the details of all evaluation benchmarks. 

\subsection{A Novel Multi-turn Benchmark} 
\subsubsection{Requirements} \label{sec:lowre-requirements}

We formulate the following desiderata for a human-evaluation benchmark considering the XLR use-case.

\textbf{1) Questions should cover real-life usage scenarios to reflect real-life usefulness.} The easiest and most likely way for speakers of low-resource Finno-Ugric languages to benefit from LLMs is through interaction via a chat-like interface. Our novel Finno-Ugric benchmark is designed to cover such real-life use cases. Consequently, it should consist of user prompts similar to real-life queries. Another benefit of using real-life data is that it helps quickly reveal the model's usefulness and potential weaknesses in practical scenarios, which standard NLP benchmarks typically do not cover.

\textbf{2) Questions should be challenging enough for LLMs to differentiate the models accurately}. \citet{zheng2023lmsys} show that challenging prompts from real-life conversations reveal larger performance gaps between different models compared to a manually designed benchmark of high-quality challenging questions.

\textbf{3) Answering questions should not require expert knowledge.} A key requirement for the benchmark is that it should comprise questions that are challenging for language models. However, such questions are often challenging for humans as well, requiring expert-level knowledge in various domains. For example, \citet{zheng2024judging} uses graduate students as labelers, considering them more knowledgeable than average crowd workers. Requiring expert-level knowledge from evaluators shrinks the potential evaluator pool, making it nearly impossible to find them from the communities of XLR language speakers.

\textbf{4) Translating the benchmark into a new language should be feasible in terms of time and content} (i.e., should not require expert knowledge). Since no data on human interactions with chat LLMs exists for XLR languages, we collect the data in English and translate it. Given the limited availability of translators, we carefully select examples that are straightforward to translate.

\subsubsection{Constructing the Dataset}

We manually collect the initial dataset from LMSYS-Chat-1M \citep{zheng2023lmsys}, which consists of real-world user interactions with LLMs. First, we extract all two-turn English conversations that have not been redacted or flagged by OpenAI moderation API. We only allow conversations with user prompts no longer than 50 tokens to ease the translation process. We then use \texttt{all-MiniLM-L12-v2} model from SentenceTransformers \citep{reimers-2019-sentence-bert} to compute the sentence embedding and apply fast clustering implemented in \texttt{sentence-transformers} which finds local groups of texts that are highly similar. We manually examine a few examples from each cluster and pick user prompts that fill the criteria specified in Chapter \ref{sec:lowre-requirements}. Finally, we removed the observed clusters from the dataset and recluster the remaining examples with a smaller similarity threshold until we had collected 248 multi-turn conversations in total.

We organize conversations into four categories: math, reasoning, writing, and general. As we wanted similarly to \citet{zheng2024judging}, the final dataset to consist of 80 questions  --- 20 from each category (potentially with follow-ups) --- the initial dataset had to be filtered. For that purpose, we asked GPT-4 to rate the difficulty of each question as was done by \citet{zheng2023lmsys}. However, we observed no correlation between the difficulty of the question and the quality of the answer given by ChatGPT when quality was assessed by GPT-4. Therefore, the final dataset was also picked manually by removing near duplicate questions and --- after looking at the generated answers --- also questions where judging the answer still seemed to require too specific knowledge. The final dataset consists of 80 real-life prompts among which 42 are multiturn. It was then translated into Võro, Komi, and Livonian by fluent speakers with a linguistic background or previous experience in translation. The translators were asked to preserve any informality of the text in the translations, e.g. missing uppercase and punctuation.   

Instead of human translators, one could use a machine translation system or a proprietary LM for test data generation or translation. We explore both options and find that proprietary models struggle to generate text in our target languages, while even the best machine translation systems produce translations that are often judged inferior to human translations (see Appendix \ref{sec:mt-bench-qe} for further details).

\begin{table*}[!ht]\centering
\scriptsize
\addtolength{\tabcolsep}{-2.8pt}
\begin{tabular}{@{}lcccccccccccccc@{}}\toprule
\multirow{3}{*}{\textbf{Model}} &\multicolumn{3}{c}{\textbf{SIB-SMUGRI}} & &\multicolumn{3}{c}{\textbf{BELEBELE-SMUGRI}} & &\multicolumn{6}{c}{\textbf{FLORES-SMUGRI}}  \\
 &\multicolumn{3}{c}{5-shot, acc} & &\multicolumn{3}{c}{3-shot, acc} & &\multicolumn{6}{c}{5-shot, BLEU} \\
\cmidrule{2-4}\cmidrule{6-8}\cmidrule{10-15}
&\textbf{VRO} &\textbf{LIV} &\textbf{KPV} &\textbf{} &\textbf{VRO} &\textbf{LIV} &\textbf{KPV} &\textbf{} &\textbf{ET-VRO} &\textbf{ET-LIV} &\textbf{RU-KPV} &\textbf{VRO-ET} &\textbf{LIV-ET} &\textbf{KPV-RU} \\\midrule
Llammas-base &78.4 \tiny{(3.7)} &69.6 \tiny{(4.1)} &64.0 \tiny{(4.3)} & &30.7 \tiny{(4.1)} &28.4 \tiny{(4.0)} &32.3 \tiny{(4.2)} & & 11.5 \tiny{(0.9)} & 4.3 \tiny{(0.5)} &1.7 \tiny{(0.4)} &28.7 \tiny{(1.5)}& 8.0 \tiny{(0.8)}&2.2 \tiny{(0.3)} \\
Llama-2-7B &57.6 \tiny{(4.4)} &60.0 \tiny{(4.4)} &58.4 \tiny{(4.4)} & &29.1 \tiny{(4.1)} &\textbf{29.9} \tiny{(4.1)} &\textbf{36.2} \tiny{(4.3)} & &11.1 \tiny{(1.0)} &4.6 \tiny{(0.6)} &1.5 \tiny{(0.3)}&11.3 \tiny{(0.9)}& 4.4 \tiny{(0.6)} &2.4 \tiny{(0.3)} \\
\midrule
\multicolumn{3}{@{}l}{\textbf{Llama-SMUGRI} (ours)} \\
Stage 1 &80.8 \tiny{(3.5)} &\textbf{75.2} \tiny{(3.9)} &65.6 \tiny{(4.3)} & &32.3 \tiny{(4.2)} &26.8 \tiny{(3.9)} &26.0 \tiny{(3.9)} & &11.5 \tiny{(1.0)} &4.2 \tiny{(0.5)} &2.6 \tiny{(0.6)}&29.6 \tiny{(1.4)} &7.2 \tiny{(0.7)}&4.1 \tiny{(0.7)}  \\
Stage 2 &78.4 \tiny{(3.7)} &65.6 \tiny{(4.3)} &74.4 \tiny{(3.9)} & &31.5 \tiny{(4.1)} &26.0 \tiny{(3.9)} &28.4 \tiny{(4.0)} & &26.5 \tiny{(1.1)} &3.4 \tiny{(0.4)} & 15.7 \tiny{(1.0)}&45.3 \tiny{(1.5)}&10.6 \tiny{(0.9)}&18.6 \tiny{(0.9)} \\
Stage 2 + parallel &\textbf{84.0} \tiny{(3.3)} &66.4 \tiny{(4.2)} &\textbf{76.8} \tiny{(3.8)} & & \textbf{35.4} \tiny{(4.3)} & 27.6 \tiny{(4.0)} &29.1 \tiny{(4.1)} & &\textbf{29.1} \tiny{(1.2)} & \textbf{4.3} \tiny{(0.5)} & \textbf{16.0} \tiny{(1.0)}&\textbf{48.7} \tiny{(1.4)}&\textbf{17.6} \tiny{(1.0)}&\textbf{22.1} \tiny{(1.3)} \\
\bottomrule
\end{tabular}
\caption{\textbf{Pre-training results for extremely low-resource Finno-Ugric languages.} Standard errors are reported for the scores in parentheses: \textit{score} (\textit{stderr}). For comparison, we report Llama-2 7B and Llammas-base \cite{kuulmets2024teaching}.
\textbf{Stage 1} consists of pre-training with high-resource related languages while \textbf{Stage 2} additionally includes Võro, Livonian and Komi. \textbf{Stage 2 + parallel} incorporates additional parallel translation data into training.}\label{tab:pretraining-results}
\end{table*}

\begin{table}[!ht]\centering
\scriptsize
\begin{tabular}{@{}lccc@{}}\toprule
\multirow{2}{*}{\textbf{Model}} &\multicolumn{3}{c}{\textbf{byte-PPL}} \\\cmidrule{2-4}
&\textbf{VRO} &\textbf{LIV} &\textbf{KPV} \\\midrule
Llammas-base &3.3548 &12.1081 &3.1959 \\
Llama-2-7B &6.1528 &14.8055 &3.1198 \\
\midrule
\multicolumn{3}{@{}l}{\textbf{Llama-SMUGRI} (ours)} \\
Stage 1 & 3.4895 &11.4210 &3.1341 \\
Stage 2 & 2.1885 &3.8351 &1.4055 \\
Stage 2 + parallel & \textbf{2.1837} &\textbf{3.7615} &\textbf{1.4050} \\
\bottomrule
\end{tabular}

\caption{\textbf{Pre-training perplexity for extremely low-resource Finno-Ugric languages.} For comparison, we report Llama-2 7B and Llammas-base \cite{kuulmets2024teaching}.}\label{tab:pretraining-results-ppl}

\end{table}

\section{Results}
\subsection{Pre-training}

\textbf{Stage 1} continued pre-training on high-resource supporting languages shows notable improvements in SIB-SMUGRI for Võro and Livonian compared to Llama-2-7B (see Stage 1 in Table~\ref{tab:pretraining-results}). There is also a clear reduction in perplexity (see Table~\ref{tab:pretraining-results-ppl}) for Võro and Livonian, while no such improvement is observed for Komi. Similarly, the Llama model fine-tuned for Estonian (Llammas-base) exhibits lower perplexity than Llama for Livonian and Võro, which are closely related to Estonian. The lack of improvement for Komi may result from its more distant relationship (see Appendix~\ref{sec:layer-representations}) to the other Finno-Ugric languages in the dataset, as well as its use of a different script. These results suggest that related languages generally improve benchmark scores for XLR languages that were not included in the training. For Belebele-SMUGRI, there is no improvement compared to Llama-2-7B, while FLORES-SMUGRI shows improvement only when translating from the XLR languages into the higher-resourced languages.

\textbf{Stage 2} pre-training, which targets XLR Finno-Ugric languages, further enhances both perplexity and, with the exception of ET-LIV translation, FLORES-SMUGRI scores, indicating that the model has acquired generative capabilities for these languages. The performance improvements on the SIB-SMUGRI benchmark are modest for Komi, while Livonian and Võro show a slight decrease from the previous stage. This stage of training has also failed to improve the Belebele scores.

\textbf{Stage 2 + parallel data} results in minimal improvements in benchmarks and perplexity, with the exception of translation tasks from the XLR languages showing larger gains. This indicates that the inclusion of parallel data has a limited impact or that our benchmarks are insufficiently sensitive to capture these effects. Nevertheless, due to the slightly positive influence observed, we will use this setup as the basis for subsequent instruction tuning.

\textbf{Benchmarks.} The current benchmarks may not effectively differentiate between models at this stage, as their small size and high standard errors limit our ability to draw fine-grained conclusions about training strategies. Additionally, the low scores on the Belebele benchmark suggest it may be too challenging for the models. In contrast, the relatively high scores on the SIB-200 benchmark could result from its simplicity, allowing the models to classify texts based on single-word clues rather than a deeper understanding of the language. Designing automatic benchmarks with an appropriate difficulty level and relevant context for these languages is an important challenge for future research.

\begin{table*}[!htp]\centering
\scriptsize
\begin{tabular}{lrrrrrrrr}\toprule
\multirow{3}{*}{\textbf{Model}} &\multicolumn{3}{c}{\textbf{BELEBELE-SMUGRI}} & &\multicolumn{3}{c}{\textbf{SIB-SMUGRI}} \\
&\multicolumn{3}{c}{0-shot, acc} & &\multicolumn{3}{c}{5-shot, acc} \\\cmidrule{2-4}\cmidrule{6-8}
&\textbf{VRO} &\textbf{LIV} &\textbf{KPV} &\textbf{} &\textbf{VRO} &\textbf{LIV} &\textbf{KPV} \\\midrule
GPT-3.5-turbo &45.7 \tiny{(4.4)} &37.8 \tiny{(4.3)} &34.6 \tiny{(4.2)} & &81.6 \tiny{(3.5)} &73.6 \tiny{(4.0)} &68.8 \tiny{(4.2)} \\
GPT-4-turbo &\textbf{70.1} \tiny{(4.1)} &40.2 \tiny{(4.3)} &\textbf{44.1} \tiny{(4.4)} &\textbf{} &\textbf{92.0} \tiny{(2.5)} &72.0 \tiny{(4.0)} &67.2 \tiny{(4.2)} \\
Llammas \cite{kuulmets2024teaching} &36.2 \tiny{(4.3)} &32.3 \tiny{(4.2)} &27.6 \tiny{(4.0)} & &80.8 \tiny{(3.5)} &78.4 \tiny{(3.7)} &63.2 \tiny{(4.3)} \\\midrule
\multicolumn{3}{l}{\textbf{Llama-SMUGRI-Instruct}} & \\
\texttt{SupInst} &42.5 \tiny{(4.4)} &30.7 \tiny{(4.1)} &\textbf{44.1} \tiny{(4.4)} & &86.4 \tiny{(3.1)} &79.2 \tiny{(3.6)} &\textbf{88.8} \tiny{(2.8)} \\
\texttt{SupInst}+\texttt{LLMTrAlpaca} &39.4 \tiny{(4.3)} &35.4 \tiny{(4.3)} &42.5 \tiny{(4.4)} & &85.6 \tiny{(3.1)} &\textbf{81.6} \tiny{(3.5)} &84.8 \tiny{(3.2)} \\
\texttt{SupInst}+\texttt{TrAlpaca} &35.4 \tiny{(4.2)} &32.3 \tiny{(4.2)} &40.2 \tiny{(4.3)} & &85.6 \tiny{(3.1)} &79.2 \tiny{(3.6)} &85.6 \tiny{(3.1)} \\
\texttt{SupInst}+\texttt{LLMTrAlpaca}+\texttt{TrInst} &44.9 \tiny{(4.4)} &\textbf{40.9} \tiny{(4.4)} &\textbf{44.1} \tiny{(4.4)} &\textbf{} &86.4 \tiny{(3.1)} &76.0 \tiny{(3.8)} &78.4 \tiny{(3.7)} \\
\texttt{SupInst}+\texttt{TrAlpaca}+\texttt{TrInst}\ &45.7 \tiny{(4.4)} &32.3 \tiny{(4.2)} &\textbf{44.1} \tiny{(4.4)} & &86.4 \tiny{(3.1)} &78.4 \tiny{(3.7)} &78.4 \tiny{(3.7)} \\
\bottomrule
\end{tabular}
\caption{\textbf{Instruction-tuning evaluation results.} Standard errors are reported for the scores: \textit{score} (\textit{stderr}).}\label{tab:instruction-tuning-results}
\end{table*}

\subsection{Instruction-Tuned Models}
Examining the scores of commercial systems in Table~\ref{tab:instruction-tuning-results}, we observe that these models exhibit at least some understanding of Võro, Livonian, and Komi. Based on benchmark scores, they seem to understand Võro and Livonian better than Komi. This could be explained by the linguistic similarity between these languages and Estonian -- an average Estonian speaker can understand most of a Võro text and some of a Livonian text, but not much Komi, as it is more distantly related and written in a different script. The performance of GPT-4-turbo and GPT-3.5-turbo aligns with this trend, with scores typically following this order. For instance, GPT-4-turbo achieves 92\% accuracy on the Belebele Estonian benchmark, so it is unsurprising that it also performs well on Võro.

Our models demonstrate comparable performance to GPT-3.5-turbo on Võro and Livonian and slightly outperform it on Komi. However, GPT-4-turbo significantly surpasses our models on Võro and matches our performance on Livonian and Komi. A similar trend emerges in the SIB benchmark: our models outperform GPT-4-turbo on Livonian and Komi but underperform on Võro. Meanwhile, GPT-3.5-turbo consistently scores lower across all XLR languages.

We observe that the different instruction-tuning strategies produce similar results. Given the small size of our benchmarks and the associated high standard errors, we cannot make definitive conclusions about which strategy is superior.

\textbf{LLM-translated instructions.} Automatic metrics indicate that instructions translated using our translation-tuned LLM achieve results comparable to those produced by the external system Neurotõlge. However, the results do not provide enough clarity or confidence to definitively favor one method over the other. These findings suggest that even in the absence of external translation systems, the translation-tuned LLM can serve as a viable alternative.

\textbf{Translation instructions.}
Incorporating a small set of translation instructions (250 for each Võro, Komi, and Livonian direction) does not lead to clear and consistent improvements in the discriminative benchmarks (see Table~\ref{tab:instruction-tuning-results}). Human evaluation of conversations (Section~\ref{sec:human-eval}) produces similar findings. However, there is a notable improvement in the translation benchmark, even with this minimal data (see Table~\ref{tab:translation-results}). We believe these translation examples mainly help the model respond in the correct language, while the underlying language and translation capabilities are already present in the base model.

\textbf{Translation evaluation.} When assessing language generation using the FLORES translation benchmark, the results in Table~\ref{tab:translation-results} show that GPT-family models can translate from Estonian to Võro quite effectively, suggesting that Võro might have been included in their training data. In contrast, the low BLEU scores for Livonian and Komi indicate very limited translation capabilities. Our LLMs, which were not exposed to translation examples during instruction-tuning, struggle to translate into Võro, Livonian, and Komi. However, they perform better in the reverse direction, even surpassing GPT models for Komi. A closer look reveals that they copy the high-resource language sentences to the output when translating to the low-resource languages. When the translated Alpaca instructions (\texttt{TrAlpaca} and \texttt{LLMTrAlpaca}) were added, we observed that when asked to translate from the low-resource languages, the models often copied the source text to the output as well, resulting in lower BLEU scores. This can be addressed by including a small amount of translation data during instruction-tuning (\texttt{TrInst}).

\subsection{Translation-tuning}

\begin{table}[!htp]\centering
\scriptsize
\addtolength{\tabcolsep}{-3pt}

\begin{tabular}{@{}lccccccccc}\toprule
\multirow{2}{*}{\textbf{Model}} &\multicolumn{2}{c}{\textbf{ET-VRO}} & &\multicolumn{2}{c}{\textbf{ET-LIV}} & &\multicolumn{2}{c}{\textbf{RU-KPV}} \\\cmidrule{2-3}\cmidrule{5-6}\cmidrule{8-9}
&$\leftarrow$ &$\rightarrow$ & &$\leftarrow$ &$\rightarrow$& &$\leftarrow$ &$\rightarrow$ \\\midrule
GPT-3.5-turbo &34.0 &15.1 & &7.7 &2.7 & &6.7 &0.5 \\
GPT-4-turbo &47.5 &20.5 & &9.9 &3.7 & &8.7 &3.1 \\\midrule
Neurotõlge &48.5 &21.2 & &\textbf{29.7} &\textbf{10.2} & &\textbf{31.5} &\textbf{17.7} \\\midrule
\multicolumn{3}{l}{\textbf{Llama-SMUGRI-Instruct}} & \\
\texttt{SupInst} &41.9 &10.7 & &11.1 &4.6 & &21.4 &3.0 \\
\texttt{SupInst}+\texttt{TrAlpaca} &16.8 &10.6 & &9.7 &4.7 & &17 &3.2 \\

\texttt{SupInst}+\texttt{LLMTrAlpaca} &23 &10.8 & &9.2 &4.6 & &13.5 &2.9 \\
\texttt{SupInst}+\texttt{TrAlpaca}+\texttt{TrInst} &45.3 &19.1 & &19.9 &5.5 & &21.4 &15.2 \\
\texttt{SupInst}+\texttt{LLMTrAlpaca}+\texttt{TrInst} &47.7 &21.2 & &20.6 &6.2 & &20.9 &16.4 \\
\midrule
\textbf{Llama-SMUGRI-translate} & \textbf{50.5} &\textbf{29.2} & &24.0 &10.0 & &23.4 &17.3 \\
\bottomrule
\end{tabular}
\caption{BLEU scores on FLORES-SMUGRI (0-shot). Translations are generated with beam size 4 for our models.}\label{tab:translation-results}
\end{table}

\begin{figure*}[h!]
    \centering
    \begin{tikzpicture}
    \begin{groupplot}[
      group style={
        group size=2 by 1,
        horizontal sep=1.25cm,
        group name=plots,
      },
        ybar,
        /pgf/bar width={5pt}, 
        width=0.5\linewidth,
        height= 0.24\linewidth,
        x axis line style = {opacity=0},
        y axis line style = {opacity=0},
        ymajorgrids = true,
        ylabel near ticks,
        ytick style={draw=none},
        legend cell align=left,
        symbolic x coords={est, vro, liv, kpv},
        xtick=data,
        enlarge x limits=0.15,
        ymax=4,
        ymin=0,
        ylabel={Score},
        xticklabels = {
            EST,
            VRO,
            LIV,
            KPV
        },
        major x tick style = transparent,
        axis y line = left,
        xticklabel style={align=center, font=\tiny\sansmath\sffamily,color=tartunlp_black},
        ylabel style={font=\small\sansmath\sffamily,color=tartunlp_black},
        title style = {font=\scriptsize\sansmath\sffamily, color=tartunlp_black},
        yticklabel style = {font=\tiny\sansmath\sffamily, color=tartunlp_black},
        label style={font=\tiny\sansmath\sffamily, color=tartunlp_black},
    ]
        \nextgroupplot[
      legend entries={ChatGPT, TrAlpaca, LLMTrAlpaca+TrInst, TrAlpaca+TrInst},
      legend to name=thelegend,
      title={NATURALNESS},
      ]
           \addplot[
                    tartunlp_black,
                    fill=tartunlp_gray, 
                    draw=none, 
                    mark=none,
                    error bars/.cd,
                    y dir=both, 
                    y explicit
                ] table[x=Group,y=Model1,y error=Model1E,col sep=comma] {data1.csv};
                \addplot[
                    tartunlp_black,
                    fill=tartunlp_red, 
                    draw=none, 
                    mark=none,
                    error bars/.cd,
                    y dir=both, 
                    y explicit
                ] table[x=Group,y=Model2,y error=Model2E,col sep=comma] {data1.csv};
                \addplot[
                    tartunlp_black,
                    fill=tartunlp_yellow, 
                    draw=none, 
                    mark=none,
                    error bars/.cd,
                    y dir=both, 
                    y explicit
                ] table[x=Group,y=Model3,y error=Model3E,col sep=comma] {data1.csv};
                \addplot[
                    tartunlp_black,
                    fill=tartunlp_violet, 
                    draw=none, 
                    mark=none,
                    error bars/.cd,
                    y dir=both, 
                    y explicit
                ] table[x=Group,y=Model4,y error=Model4E,col sep=comma] {data1.csv};
    
        \nextgroupplot[
      legend entries={ChatGPT, TrAlpaca, LLMTrAlpaca+TrInst, TrAlpaca+TrInst},
      legend to name=thelegend,
      title={HELPFULNESS},
      legend style={
                    font=\tiny\sansmath\sffamily,
                    color=tartunlp_black,
                    column sep=0.5ex,
                    draw=none,
                    draw opacity=0,
                    fill=none,
                    legend columns = -1,
                    /tikz/every even column/.append style={column sep=0.5cm},
            }
    ]
            \addplot[
                    tartunlp_black,
                    fill=tartunlp_gray, 
                    draw=none, 
                    mark=none,
                    error bars/.cd,
                    y dir=both, 
                    y explicit
                ] table[x=Group,y=Model1,y error=Model1E,col sep=comma] {data2.csv};
                \addplot[
                    tartunlp_black,
                    fill=tartunlp_red, 
                    draw=none, 
                    mark=none,
                    error bars/.cd,
                    y dir=both, 
                    y explicit
                ] table[x=Group,y=Model2,y error=Model2E,col sep=comma] {data2.csv};
                \addplot[
                    tartunlp_black,
                    fill=tartunlp_yellow, 
                    draw=none, 
                    mark=none,
                    error bars/.cd,
                    y dir=both, 
                    y explicit
                ] table[x=Group,y=Model3,y error=Model3E,col sep=comma] {data2.csv};
                \addplot[
                    tartunlp_black,
                    fill=tartunlp_violet, 
                    draw=none, 
                    mark=none,
                    error bars/.cd,
                    y dir=both, 
                    y explicit
                ] table[x=Group,y=Model4,y error=Model4E,col sep=comma] {data2.csv};
                 
            \legend{ChatGPT, TrAlpaca, LLMTrAlpaca+TrInst, TrAlpaca+TrInst}
            
            \coordinate (c2) at (rel axis cs:1,1);
        \end{groupplot}
        \coordinate (c3) at ($(rel axis cs:0,1)!.5!(c2)$);

        \node[below] at (c3 |- current bounding box.south)
          {\pgfplotslegendfromname{thelegend}};

    \end{tikzpicture}
    \caption{Human evaluation scores for naturalness and helpfulness across different models.}
\end{figure*}

\begin{figure*}[h]
    \centering
        \centering
        \includegraphics[width=0.95\textwidth]{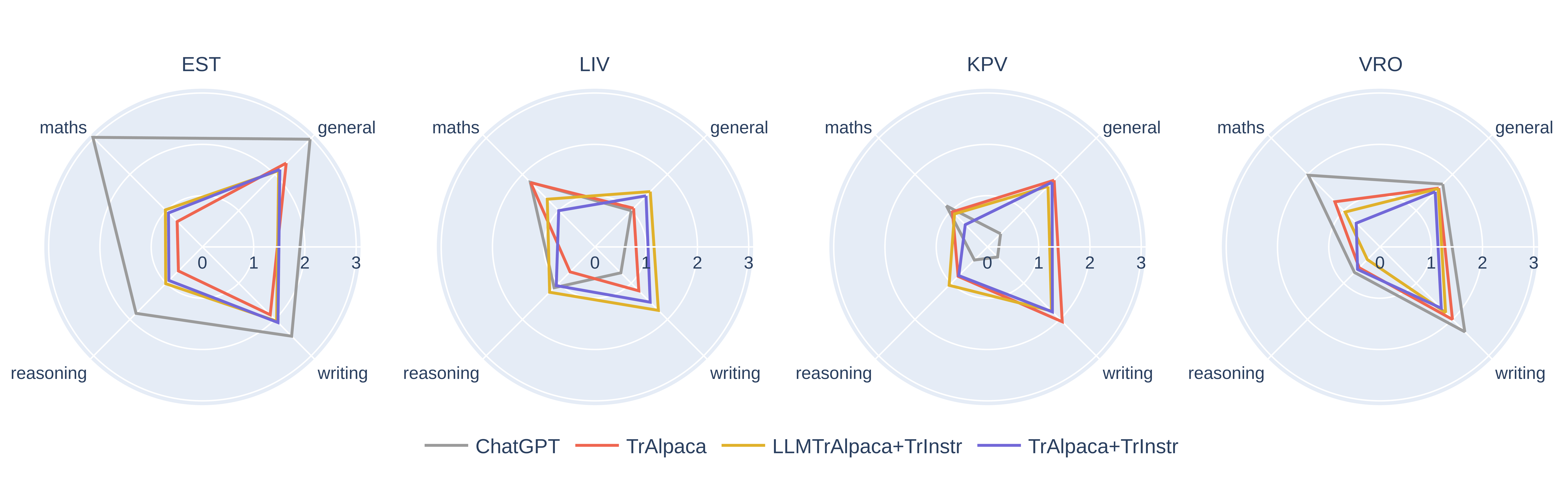}
        \label{fig:image3}
    \caption{Helpfulness across languages and categories.}
    \label{fig:usefulness-by-category}
\end{figure*}

We compare our LLM-based translation models with Neurotõlge, which supports low-resource Finno-Ugric languages. Our translation-tuned models outperform Neurotõlge in both the VRO-ET and ET-VRO translation directions (see Table~\ref{tab:translation-results}). For the ET-LIV and RU-KPV directions, our models achieve performance that is comparable to Neurotõlge. However, in translations from low-resource to high-resource languages -- except for Võro -- our models do not perform as well.

\subsection{Human Evaluation}\label{sec:human-eval}
We select 3 instruction-tuned models for human evaluation: \texttt{TrAlpaca}, \texttt{LLMTrAlpaca}+\texttt{TrInst} and \texttt{TrAlpaca}+\texttt{TrInst}. As a baseline, we use GPT-3.5-turbo, which was freely accessible via a chat interface\footnote{\href{https://chatgpt.com/}{https://chatgpt.com/}} at the time of the evaluation. For each target language, we designed a survey where participants rate the helpfulness of the answer from a randomly selected model on a 5-point Likert scale. Additionally, we ask participants to rate how natural the response sounds in the target language, as \citet{kuulmets2024teaching} notes that model outputs often sound unnatural in these languages. The surveys were distributed within target language communities via social media and direct outreach to speakers (see Appendix \ref{app:survey} for the screenshot of the survey). We did not collect any personal data from respondents.

In addition to Võro, Livonian, and Komi, we also gather and present human evaluation data for Estonian, which is closely related to both Võro and Livonian.  At the same time, Estonian is well-supported by GPT-3.5-turbo \citep{kuulmets2024teaching}, providing a meaningful anchor point for comparing our human evaluation results.

The results indicate that our models underperform in helpfulness compared to GPT-3.5-turbo in Estonian, which aligns with previous findings \cite{kuulmets2024teaching}. A similar disparity exists for Võro, where our models still lag behind. However, for both Võro and Livonian, the helpfulness scores of our models are comparable to those of GPT-3.5-turbo. In contrast, our system outperforms the commercial baseline for Komi. While variations in annotator expectations may influence results across different languages, it is noteworthy that our models consistently achieve similar helpfulness scores across various languages.

Comparisons by category (see Figure~\ref{fig:usefulness-by-category}) reveal that the scores for GPT-3.5-turbo are inflated by examples in the \textit{maths} and \textit{reasoning} categories, where our models demonstrate less helpfulness. In contrast, our models perform comparably in the \textit{general} and \textit{writing} categories. Notably, in Komi, our models surpass GPT-3.5-turbo in both \textit{general} and \textit{writing} tasks, while achieving similar scores in the \textit{maths} and \textit{reasoning} tasks.

In terms of response naturalness, GPT-3.5-turbo performs slightly better for Estonian; however, our models demonstrate greater naturalness in all other languages, especially in Komi, where the difference is particularly pronounced.

When comparing our trained models, no clear ranking emerges, reinforcing the findings from automatic benchmarks that incorporating translation instructions does not produce significant advantages. Additionally, there is little difference between using LLM-translated instructions and those translated by an external system.

\section{Conclusion}
We implemented a comprehensive approach encompassing data collection, instruction tuning, and human evaluation for three extremely low-resource Finno-Ugric languages: Võro, Livonian, and Komi. Our contributions include an exploration of pre-training and instruction-tuning strategies, leading to the development of open-source multilingual base and instruction-tuned models for these languages. We also extend the automatic evaluation benchmarks, Belebele and SIB-200, to include Komi, Livonian, and Võro, and we introduce a novel multi-turn conversational benchmark, \textsc{smugri-MT-bench}. Human evaluation using \textsc{smugri-MT-bench} demonstrates that our models surpass \texttt{GPT-3.5-turbo} in terms of naturalness and achieve higher helpfulness for Komi, while maintaining comparable levels for the other low-resource languages.

\section*{Limitations}\label{sec:limitations}
There are several limitations that may affect the robustness and generalizability of our findings. Firstly, the automatic benchmarks used are small and exhibit high standard errors, making fine-grained comparisons difficult. This issue is compounded by our reliance on the FLORES-200 dataset, which limits the scope of our evaluation to the specific topics and set of sentences it covers. Furthermore, our automatic evaluation utilized only three tasks, which constrains the comprehensiveness of our assessment. From these three, only one (translation) measured generative performance, as no other suitable benchmarks exist for these languages. This narrow focus on translation might not fully capture the generative capabilities of the models across different tasks. However, human evaluation addresses these concerns to some extent, providing a more detailed and reliable assessment of the model's quality in a multi-turn chat assistant scenario.

The heavy reliance on the FLORES-200 dataset is caused by the difficulties related to creating new datasets. Creating high-quality benchmarks for XLR languages is tricky because the data can not be obtained by machine translating benchmarks from other languages, as the machine translation systems are potentially too weak. Additionally, hiring professional translators is difficult due to the scarcity or absence of individuals experienced in translating these languages, particularly when the languages are not officially recognized. Finally, since finding human annotators for XLR languages in itself is challenging, finding expert-level annotators becomes almost impossible, and thus, the set of prompts used for human evaluation must be constructed so that assessing the quality of the answer would not require any specific expert-level knowledge.

A limitation of our instruction-tuning process is that we only used machine-translated instructions for the XLR languages. As a result, some of these instructions were of low quality, potentially affecting the overall performance and reliability of the fine-tuned models.

Our emphasis on Finno-Ugric languages means that our findings might not apply to other language families, which could present different challenges or yield different results in a more diverse multilingual context. To address these limitations, future research should aim to develop larger and more diverse benchmarks and apply similar methodologies to a broader range of low-resource languages to validate and extend our findings.

\section*{Ethics Statement}
Our models have not been extensively tested for the generation of harmful content. Furthermore, we were unable to check the training and instruction-tuning data for harmful content due to their sheer volume. Thus, we can not guarantee the models' harmlessness and advise them to be used with this in mind only for research purposes. Furthermore, our models still make many mistakes when generating the responses, and their output should not be considered an accurate representation of the low-resource languages without manual verification.

\section*{Acknowledgments}
This work was supported by the Estonian Research Council grant PRG2006 (Language Technology for Low-Resource Finno-Ugric Languages and Dialects). All computations were performed on the LUMI Supercomputer through the University of Tartu's HPC center.

We would like to express our sincere gratitude to the translators Janek Vaab (Võro), Aleksei Ivanov (Komi), and Marili Tomingas (Livonian). We would also like to thank Võro, Komi, and Livonian communities for their help in annotating the data. 

\bibliography{custom,anthology}

\appendix

\section{Collecting Data for Human Evaluation}
\label{app:survey}
\begin{figure}
    \centering
    \includegraphics[width=1.0\linewidth]{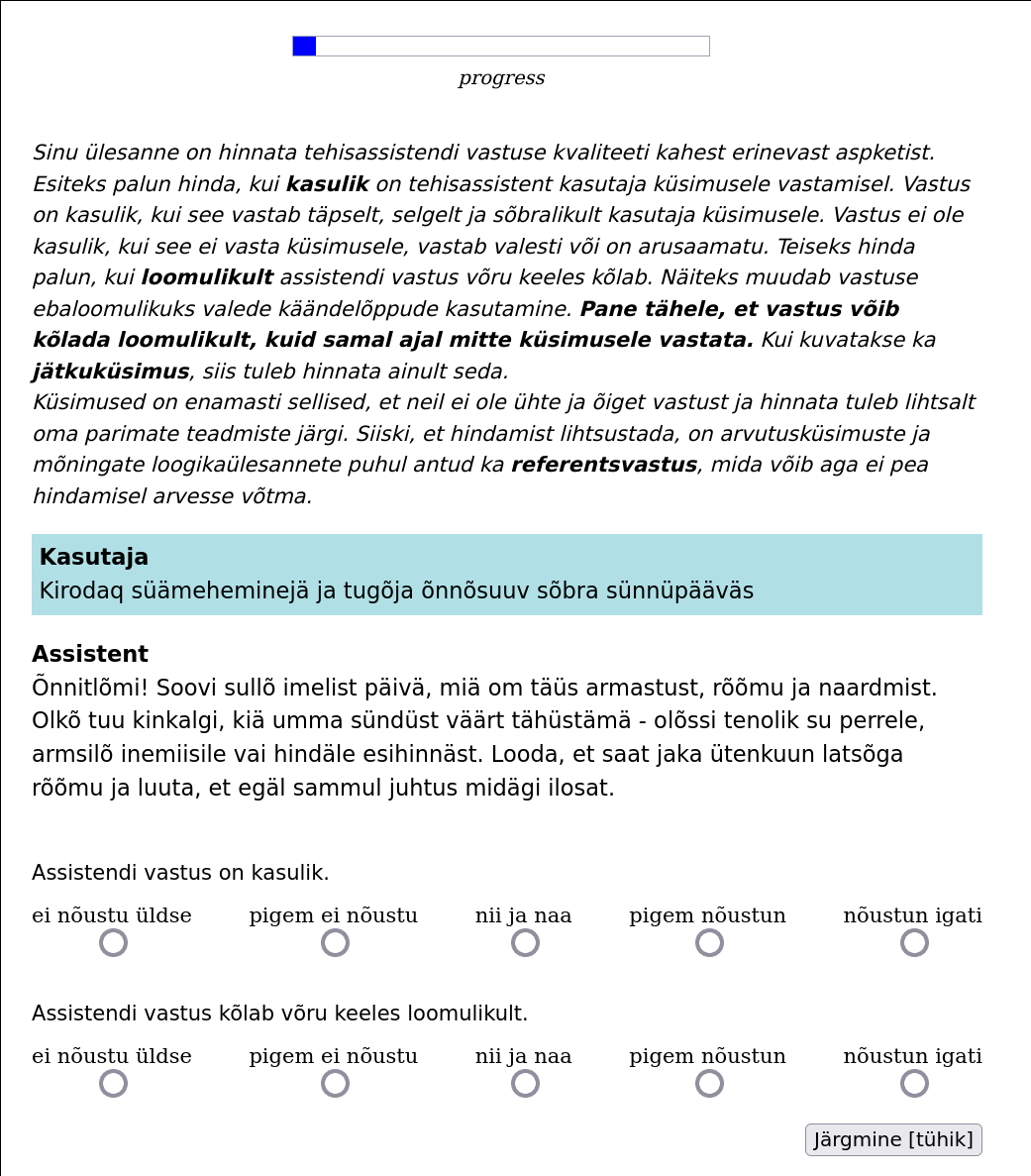}
    \caption{Screenshot of the survey that was used to collect human annotations.}
    \label{fig:survey}
\end{figure}

The screenshot of the survey is shown in Figure~\ref{fig:survey}. For Võro, Livonian, and Estonian, the instructions were given in Estonian, while for Komi, they were given in Russian.

\section{Training Details}\label{sec:training-params}
The hyperparameters of pre-training stages 1 and 2 are listed in Table~\ref{tab:params-pretraining}. The instruction-tuning and translation-tuning parameters are in Table~\ref{tab:params-finetuning}. The first epoch was used for evaluating instruction-tuned models.

All the models were trained using 4 AMD MI250x GPUs (acting as 8 units) on the LUMI supercomputer. We report GPU-hours elapsed for model training in Table~\ref{tab:gpu-hours}.

\begin{table}[!htp]\centering
\scriptsize
\begin{tabular}{lcc}\toprule
Parameter &Stage 1 &Stage 2 (translate)\\\midrule
updates &19073 & 2985 (3013) \\
LR &4.00e-5 &2.00e-5 \\
LR-schedule &\multicolumn{2}{c}{cosine decay to 10\%} \\
context length &\multicolumn{2}{c}{2048} \\
batch size &\multicolumn{2}{c}{256} \\
warmup ratio &\multicolumn{2}{c}{0.01} \\
weight decay &\multicolumn{2}{c}{0.05} \\
precision &\multicolumn{2}{c}{bfloat16} \\
optimizer &\multicolumn{2}{c}{AdamW} \\
packing &\multicolumn{2}{c}{yes} \\
\bottomrule
\end{tabular}
\caption{Pre-training hyperparameters.}\label{tab:params-pretraining}
\end{table}

\begin{table}[!htp]\centering
\scriptsize
\begin{tabular}{lrr}\toprule
Parameter &Value \\\midrule
LR &2.00e-5 \\
LR-schedule &cosine decay to 10\% \\
context length &2048 \\
batch size &256 \\
epochs &2 \\
warmup ratio &0.01 \\
weight decay &0.05 \\
precision &bfloat16 \\
optimizer &AdamW \\
packing &no \\
\bottomrule
\end{tabular}
\caption{Instruction-tuning and translation-tuning hyperparameters.}\label{tab:params-finetuning}
\end{table}

\begin{table}[!htp]\centering
\scriptsize
\begin{tabular}{lrr}\toprule
\textbf{Model} &\textbf{GPU-hours} \\\midrule
\textbf{Base:} & \\
Stage 1 &2008 \\
Stage 2 &308 \\
Stage 2 + translate &316 \\\midrule
\textbf{Instruction:} & \\
LLMTrAlpaca+TrInst &39 \\\midrule
TrTuning &39 \\
\bottomrule
\end{tabular}
\caption{GPU-hours elapsed for training the models.}\label{tab:gpu-hours}
\end{table}

\begin{table*}[!htp]\centering
\scriptsize
\begin{tabular}{lcccccccccccc}\toprule
\multirow{2}{*}{\textbf{Sampling}} &\multicolumn{3}{c}{\textbf{byte-PPL}} &\textbf{} &\multicolumn{3}{c}{\textbf{Epochs}} &\textbf{} &\multicolumn{3}{c}{\textbf{Proportion}} \\\cmidrule{2-4}\cmidrule{6-8}\cmidrule{10-12}
&\textbf{VRO} &\textbf{LIV} &\textbf{KPV} &\textbf{} &\textbf{VRO} &\textbf{LIV} &\textbf{KPV} &\textbf{} &\textbf{VRO} &\textbf{LIV} &\textbf{KPV} \\\midrule
\textbf{Unimax} \\
N=1 &2.3072 &4.1986 &1.4508 & &1 &1 &1 & &2.4\% &0.4\% &97.2\% \\
N=4 &\textbf{2.1885} &\textbf{3.8351} &\textbf{1.4055} &\textbf{} &4 &4 &2.5 & &3.7\% &0.7\% &95.6\% \\
N=8 &2.5983 &4.7250 &1.4159 & &8 &8 &2.4 & &7.5\% &3.3\% &81.0\% \\ \midrule
Proportional &2.1983 &3.8352 &1.4065 & &2.5 &2.5 &2.5 & &2.4\% &0.4\% &97.2\% \\
\bottomrule
\end{tabular}
\caption{The effect of Unimax N (max data repeat epochs) on held-out validation set byte perplexity.}\label{tab:unimax-n}

\end{table*}

\section{Choice of Unimax N}\label{sec:unimax}
We chose the Unimax N according to the byte perplexity on our held-out validation set, with the best value for our setup being 4 (see Table~\ref{tab:unimax-n}). 

\citet{muennighoff2023scaling} found that repeating data for 4 epochs is almost as good as new data with improvements continuing beyond 4 epochs for pre-training LLMs. We find that for continued pre-training with very small datasets and small data budgets, 4 epochs of repetition (Unimax N=4) provides an improvement in perplexity over 1 epoch of data for \textbf{Stage 2}. However, already at 8 epochs, the perplexity drops, suggesting overfitting (see Appendix~\ref{sec:unimax}). Thus we keep the maximum repetitions at 4 and conclude that the number of repetitions of smaller datasets should be carefully chosen to avoid over- or underfitting.

\begin{table}[h]\centering
\scriptsize
\begin{tabular}{lrrrrr}\toprule
&ET &VRO &LIV &KPV \\\midrule
surveys submitted &45 &17 &6 &27 \\
answers graded &1708 &836 &279 &1306 \\
grades per question &2.8 &1.74 &0.58 &2.7 \\
\bottomrule
\end{tabular}
\caption{Human evaluation data collection statistics.}\label{tab:human-eval-stats}
\end{table}

\section{Evaluation details}\label{sec:eval-details}
 The base models are evaluated with \texttt{lm-evaluation-harness} \citep{eval-harness}. For instruction-tuned models' SIB-SMUGRI outputs that do not conform to the expected format, we use \texttt{GPT-4-turbo} to verify that the prediction matches the ground truth. We calculate standard errors using bootstrap resampling implemented in \texttt{lm-evaluation-harness} \cite{eval-harness}. The evaluation prompts are listed in Figure~\ref{fig:eval-prompts}. For Belebele, the instruction-tuned models' zero-shot evaluation method is based on \citet{bandarkar2023belebele}.

 \texttt{GPT-4-turbo} version used in evaluations was \texttt{gpt-4-turbo-2024-04-09} and \texttt{GPT-3.5-turbo} version used was \texttt{gpt-3.5-turbo-0125}.
 
 We evaluate translations quality using BLEU \citep{papineni-etal-2002-bleu} calculated with sacreBLEU\footnote{signature: \texttt{nrefs:1|case:mixed|eff:no|tok:13a\\|smooth:exp|version:2.4.2}} \cite{post-2018-call}.

 The held-out validation set (see Table~\ref{tab:heldout-valid}) used to calculate perplexity is sampled from our pre-training data.

 \begin{table}[!htp]\centering
\scriptsize
\begin{tabular}{lrrr}\toprule
\textbf{Language} &\textbf{Characters} &\textbf{Examples} \\\midrule
LIV &86842 &1246 \\
VRO &131373 &110 \\
KPV &1308290 &500 \\
\bottomrule
\end{tabular}
\caption{Held-out validation set sizes. Examples for Livonian are sentences. For other languages they are documents.}\label{tab:heldout-valid}
\end{table}

\section{Võro Data Collection}\label{sec:vro-composition}
We collect Võro data from Võro language Wikipedia dump \cite{wikidump}, Corpus of Fiction in Võro and Seto languages\footnote{\href{https://metashare.ut.ee/repository/browse/corpus-of-fiction-in-voro-and-seto-languages/2cf454fca0d411eebb4773db10791bcf485f3f9e7dee447b983f21b074ad3835}{https://metashare.ut.ee/repository/browse/corpus-of-fiction-in-voro-and-seto-languages/2cf454fca0d411eebb47\\73db10791bcf485f3f9e7dee447b983f21b074ad3835}}, Additionally, we scraped Võro language newspaper articles from \textit{Uma Leht}\footnote{\href{https://umaleht.ee/}{https://umaleht.ee/}}. Since the Seto dialect is similar to Võro, we do not filter it out of our Võro datasets that contain it,  and additionally include "Setomaa" newspaper corpus\footnote{\href{https://metashare.ut.ee/repository/browse/setomaa-newspaper-corpus/3303e60ca0d411eebb4773db10791bcf2d01e0b55ce2419db34ef402460a1c99/}{https://metashare.ut.ee/repository/browse/setomaa-newspaper-corpus/3303e60ca0d411eebb4773db10791b\\cf2d01e0b55ce2419db34ef402460a1c99/}} which is also in Seto dialect. The collected Võro dataset composition is shown in Table~\ref{tab:vro-composition}.

\begin{table}[!htp]\centering
\scriptsize
\begin{tabular}{lrrrr}\toprule
Name &Documents &Characters &Sentences \\\midrule
\textbf{Võro} & & & \\
Wikipedia (2024.02.20) &6385 &3879212 &88550 \\
Fiction corpus &399 &1987446 &32121 \\
Umaleht (scraped) &3392 &6280588 &93958 \\\midrule
\textbf{Seto dialect} & & & \\
Fiction corpus &8 &76361 &869 \\
Setomaa corpus &397 &1791268 &20693 \\
\bottomrule
\end{tabular}
\caption{Võro data composition by source.}\label{tab:vro-composition}

\end{table}

\begin{table*}[!htp]\centering
\scriptsize
\begin{tabular}{lrrrrrrrr}\toprule
\textbf{Dataset} &\textbf{LIV} &\textbf{VRO} &\textbf{KPV} &\textbf{ET} &\textbf{FI} &\textbf{EN} &\textbf{RU} \\\midrule
Supporting language instructions:\\
Aya \cite{singh2024aya} & & & & &742 &3944 &423 \\
OASST-2 \cite{kopf2023openassistant} & & & & &5 &3514 &681 \\
FLAN-V2 \cite{pmlr-v202-longpre23a} & & & & & &5000 & \\
Alpaca-GPT-4 \cite{peng2023instruction} & & & & & &20000 & \\
Alpaca-est \cite{kuulmets2024teaching} & & & &20000 & & & \\\midrule
TrAlpaca (ours) &1000 &1000 &1000 & & & & \\\midrule
\textbf{TOTAL} &1000 &1000 &1000 &20000 &747 &32458 &1104 \\
\bottomrule
\end{tabular}
\caption{Instruction-tuning data with the number of sentences sampled}\label{tab:instruction-dataset}
\end{table*}

\section{Instruction-tuning details}\label{sec:instruction-tuning-details}

The composition of our instruction-tuning dataset is listed in Table~\ref{tab:instruction-dataset}. Instructions are formatted into a chat-format shown in Figure~\ref{tab:chat-format}. The translation-tuning data format is shown in Figure~\ref{tab:trtuning-format}. The fine-tuning loss is calculated on target (assistant) tokens while the rest of the prompt is masked.

\section{Instruction translation details}\label{sec:instruction-translation}
When using our translation-tuned models for translating instructions (for \texttt{LLMTrAlpaca}), the models sometimes leave sentences untranslated in an unpredictable manner. Consequently, we removed examples where the BLEU \cite{papineni-etal-2002-bleu} score between the original and translated text exceeds 70. This process may also eliminate some valid examples, as identical texts can occur in some cases.

In preliminary experiments, we observed that the model sometimes struggles with multi-line or multi-sentence inputs, which are essential for accurately translating instructions that often consist of entire texts from Alpaca-style examples. To address this issue, we concatenate 50\% of the training sentences into chunks of 2 to 6 sentences, training the model to handle longer inputs effectively. We refer to this configuration \texttt{TrTuningConcat} and the regular translation instructions as \texttt{TrTuning}. 

\begin{table*}[!hpt]\centering
\scriptsize
\begin{tabular}{lrrrrrrrrrrrr}\toprule
Dataset &VRO-ET &LIV-ET &LIV-LV &LIV-EN &KPV-ET &KPV-FI &KPV-RU &KPV-EN &KPV-LV &&TOTAL \\\midrule
\texttt{TrInst} &500 &500 &500 &493 &500 &500 &500 &500 &500 &&\textbf{4493} \\
\texttt{TrTuning} &28505 &14215 &11608 &493 &3876 &7273 &100000 &7288 &5020 &&\textbf{178278} \\
Stage 2 + parallel &28504 &14212 &11606 &492 &3835 &7272 &81487 &7286 &5018 &&\textbf{159712} \\
\bottomrule
\end{tabular}
\caption{\textbf{Number of sentences of parallel data used in various training configurations.} In all cases, the language pair data is split equally; for instance, in ET-LIV, 50\% of the reported sentences are for ET$\rightarrow$LIV and the remaining 50\% for LIV$\rightarrow$ET. The data is sourced from  \citet{yankovskaya-etal-2023-machine,DBLP:conf/acl/RiktersTTEF22,tars-etal-2022-teaching,tars2021extremely}.}\label{tab:parallel-sents}
\end{table*}

\begin{table}[!htp]\centering
\scriptsize
\addtolength{\tabcolsep}{-3pt}

\begin{tabular}{@{}lccccccccc}\toprule
\multirow{2}{*}{\textbf{Model}} &\multicolumn{2}{c}{\textbf{ET-VRO}} & &\multicolumn{2}{c}{\textbf{ET-LIV}} & &\multicolumn{2}{c}{\textbf{RU-KPV}} \\\cmidrule{2-3}\cmidrule{5-6}\cmidrule{8-9}
&$\leftarrow$ &$\rightarrow$ & &$\leftarrow$ &$\rightarrow$& &$\leftarrow$ &$\rightarrow$ \\
\midrule
Neurotõlge &48.5 &21.2 & &\textbf{29.7} &\textbf{10.2} & &\textbf{31.5} &\textbf{17.7} \\\midrule
\multicolumn{3}{l}{\textbf{Llama-SMUGRI-translate}} & \\
\texttt{TrTuning} &50.5 &\textbf{29.2} & &24.0 &10.0 & &23.4 &17.3 \\
\texttt{TrTuningConcat} &\textbf{51.7} &28.7 & &22.9 &9.7 & &23.5 &17.4 \\
\bottomrule
\end{tabular}
\caption{BLEU scores on FLORES-SMUGRI (0-shot). Translations are generated with beam size 4 for our models. TrTuningConcat uses concatenated sentences from multiple examples in a single translation instruction.}\label{tab:translation-results-ablation}
\end{table}

We find that this concatenation does not harm model translation quality (see Table~\ref{tab:translation-results-ablation}). Additionally, we observed more consistent outputs when translating whole instructions.

To get a glimpse into the quality of the translations, we conducted a small-scale human evaluation with native speakers or, in the case of Livonian a fluent speaker, due to the lack of native speakers. Given the original and the translation of 20 randomly chosen instructions, which stay the same across translation models, the evaluators were asked to rate the Fluency (\textit{How fluent and natural does the translation sound in the target language?}) and Consistency (\textit{Does the translation preserve the meaning and intent of the source text?}) on a 5-point Likert scale (see Table~\ref{tab:eval-guides} for evaluation guidelines). We also asked them to report if the resulting \textit{instruction-input-output} triplet does not form a correct instruction, e.g. the output does not satisfy the instruction. 

\begin{figure}[ht]
    \centering
        \centering
        \includegraphics[width=\columnwidth]{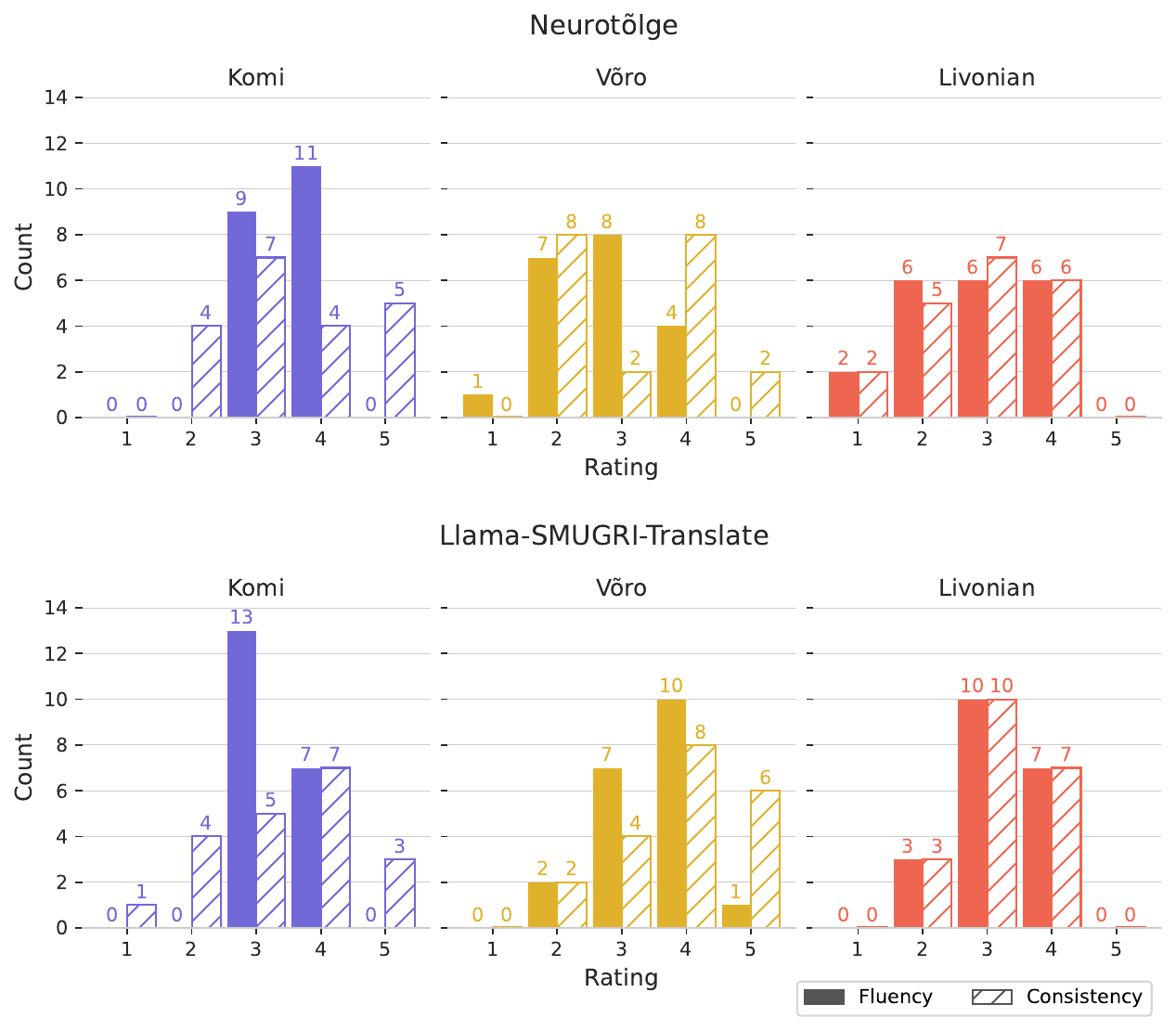}
    \caption{\textbf{Neurotõlge} and \textbf{Llama-SMUGRI-translate} instruction translation human evaluation scores.}
    \label{fig:he-histogram-combined}
\end{figure}

We report the evaluation summary in Table~\ref{tab:he-sum} for Neurotõlge and Llama-SMUGRI-translate. In general, Llama-SMUGRI-translate produces better results for Võro and Livonian, achieving higher fluency and consistency scores than Neurotõlge on average. For an acceptable instruction translation, we would like the fluency and consistency to be at least 3, and the resulting instruction should still be correct. We see that for Neurotõlge, this is only achieved for 45\% of evaluated instructions in Võro and Livonian, while for Komi 80\% of instructions are acceptable. For Llama-SMUGRI-translate, we see this condition is satisfied for 70\%, 80\%, and 65\% of translation for Komi, Võro, and Livonian, respectively. While this shows that machine translation can be a feasible option in some cases, it is far from ideal for instructions. We also report histograms of human ratings in Figure~\ref{fig:he-histogram-combined}.

\begin{table}[!htp]\centering
\scriptsize
\begin{tabular}{lp{0.8\columnwidth}}\toprule
\multicolumn{2}{c}{\textbf{Fluency}} \\
\textbf{5} &The translation is perfectly fluent, with no grammatical errors, unnatural phrasing, or awkward expressions. \\
\textbf{4} &The translation is mostly fluent, with minor grammatical or stylistic issues that do not affect readability. \\
\textbf{3} &The translation is somewhat fluent, but noticeable issues (e.g., awkward phrasing or grammatical errors) hinder smooth reading. \\
\textbf{2} &The fluency of the translation is poor, with significant issues that make it difficult to understand in parts. \\
\textbf{1} &The translation is completely unnatural or ungrammatical, making it incomprehensible. \\
& \\
\multicolumn{2}{c}{\textbf{Consistency}} \\
\textbf{5} &The translation fully preserves the meaning of the source text, with no omissions, additions, or distortions. \\
\textbf{4} &The translation preserves the overall meaning, but there are minor inaccuracies or nuances lost. \\
\textbf{3} &The translation conveys the general meaning, but there are noticeable issues (e.g., omissions or slight distortions). \\
\textbf{2} &The translation distorts the meaning significantly or omits important details, making it partially inaccurate. \\
\textbf{1} &The translation fails to convey the meaning of the source text entirely. \\
\bottomrule
\end{tabular}
\caption{Instruction translation human evaluation guidelines.}\label{tab:eval-guides}
\end{table}

\begin{table}[!htp]\centering
\scriptsize
\begin{tabular}{lccccc}\toprule
\textbf{Lang} &\textbf{Fluency} &\textbf{Consistency} &\textbf{Incorrect} &\textbf{Both $\geq 3$} \\\midrule
\multicolumn{5}{l}{\textbf{Neurotõlge}} \\\midrule
KPV &3.55 &3.5 &5\% &80\% \\
VRO &2.75 &3.2 &0\% &45\% \\
LIV &2.8 &2.85 &15\% &45\% \\\midrule
\multicolumn{5}{l}{\textbf{Llama-SMUGRI-translate}} \\\midrule
KPV &3.35 &3.35 &5\% &70\% \\
VRO &3.5 &3.9 &0\% &80\% \\
LIV &3.2 &3.2 &20\% &65\% \\
\bottomrule
\end{tabular}
\caption{Average fluency and consistency ratings of instruction translation (out of 5). Incorrect - 
 the translation does not form a correct \textit{instruction-input-response} triplet. Both $\geq 3$ - the percentage of translated instructions where the Fluency and Consistency are at least 3, and the instruction is correct.}\label{tab:he-sum}
\end{table}

\section{Parallel data}\label{sec:parallel-data}
Composition of the parallel data is shown in Table~\ref{tab:parallel-sents}.

\begin{figure}
  \centering
  \begin{minipage}{0.5\textwidth}
   \newtcolorbox{alpacabox}{
      colback=white,
      colframe=black,
      arc=4pt, 
      boxrule=0.5pt, 
      fontupper=\ttfamily, 
      left=6pt, 
      right=6pt, 
    }
    
    \begin{alpacabox}
    \begin{small}
    <|user|>\\
    Tere!\\
    <|assistant|>\\
    Tere! Kas saaksin teid kuidagi aidata?</s>\\
    <|user|>\\
    Kuidas alustada kirja kirjutamist?\\
    <|assistant|>
    \end{small}
    \end{alpacabox}
  \end{minipage}
  \caption{Chat format following \citet{wang2023how} and \citet{kuulmets2024teaching}. The model responds after \texttt{<|assistant|>}.}\label{tab:chat-format}
\end{figure}

\begin{figure}
  \centering
  \begin{minipage}{0.5\textwidth}
   \newtcolorbox{alpacabox}{
      colback=white,
      colframe=black,
      arc=4pt,
      boxrule=0.5pt,
      fontupper=\ttfamily,
      left=6pt,
      right=6pt,
    }
    
    \begin{alpacabox}
    \begin{small}
    <|system|>\\
    Translate the following \{src\_lang\} text into \{tgt\_lang\}.\\
    <|user|>\\
    \{src\_text\}\\
    <|assistant|>\\
    \{tgt\_text\}</s>
    \end{small}
    \end{alpacabox}
  \end{minipage}
  \caption{Translation-tuning data format based on Figure~\ref{tab:chat-format}.}\label{tab:trtuning-format}
\end{figure}

\section{Llama-SMUGRI Representations}\label{sec:layer-representations}

\begin{figure}[ht]
    \centering
        \centering
        \includegraphics[width=\columnwidth]{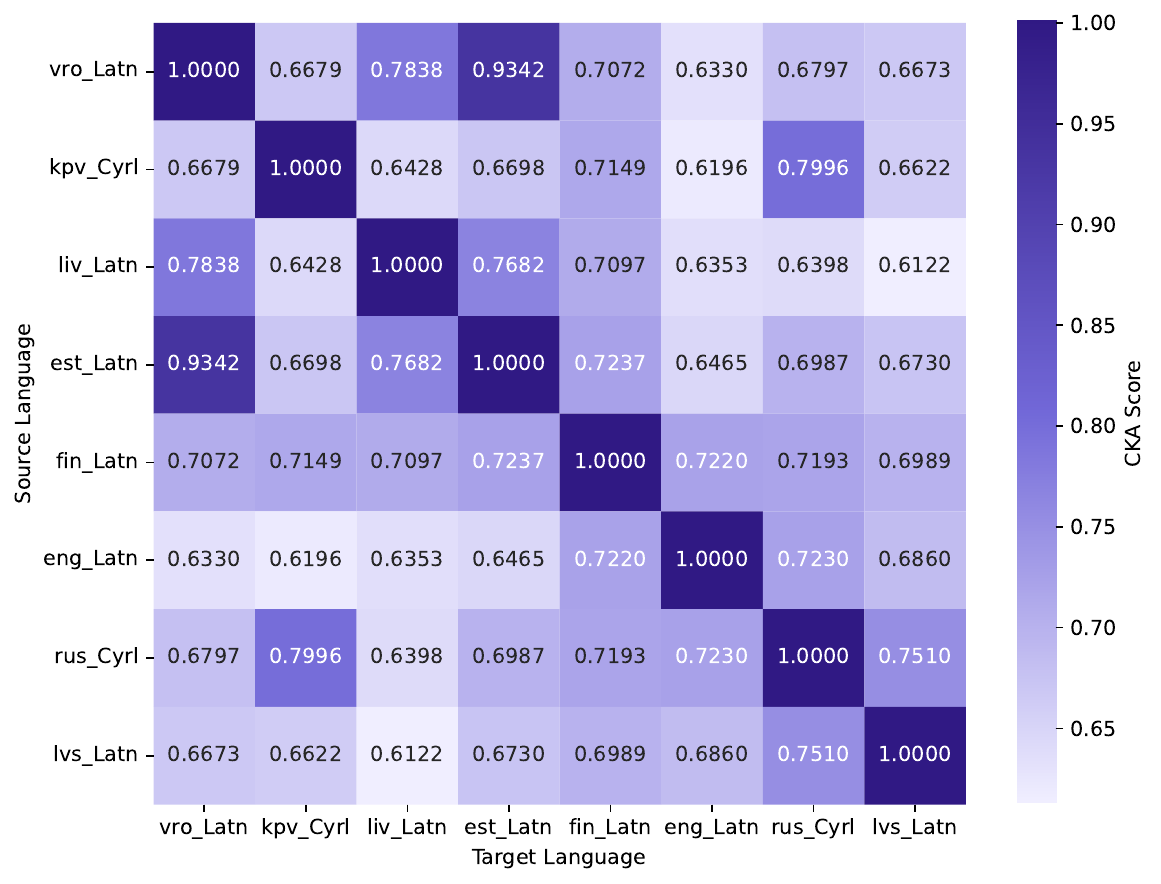}
    \caption{Llama-SMUGRI (Stage 2 + parallel) CKA scores \cite{pmlr-v97-kornblith19a} of mean-pooled layer 16 embeddings.}
    \label{fig:cka}
\end{figure}

The CKA scores \cite{pmlr-v97-kornblith19a} in Table~\ref{fig:cka} indicate that the more closely related Finno-Ugric languages -- Estonian, Võro, and Livonian -- written in the Latin script exhibit more similar representations in the intermediate layers. Meanwhile, Komi shows the highest CKA score with Russian, the only other language in our experiments that uses the Cyrillic script, while having a lower similarity score with languages for which the model has seen the most training data, such as English and Estonian.

\begin{figure*}[ht]
    \centering
        \centering
        \includegraphics[width=\textwidth]{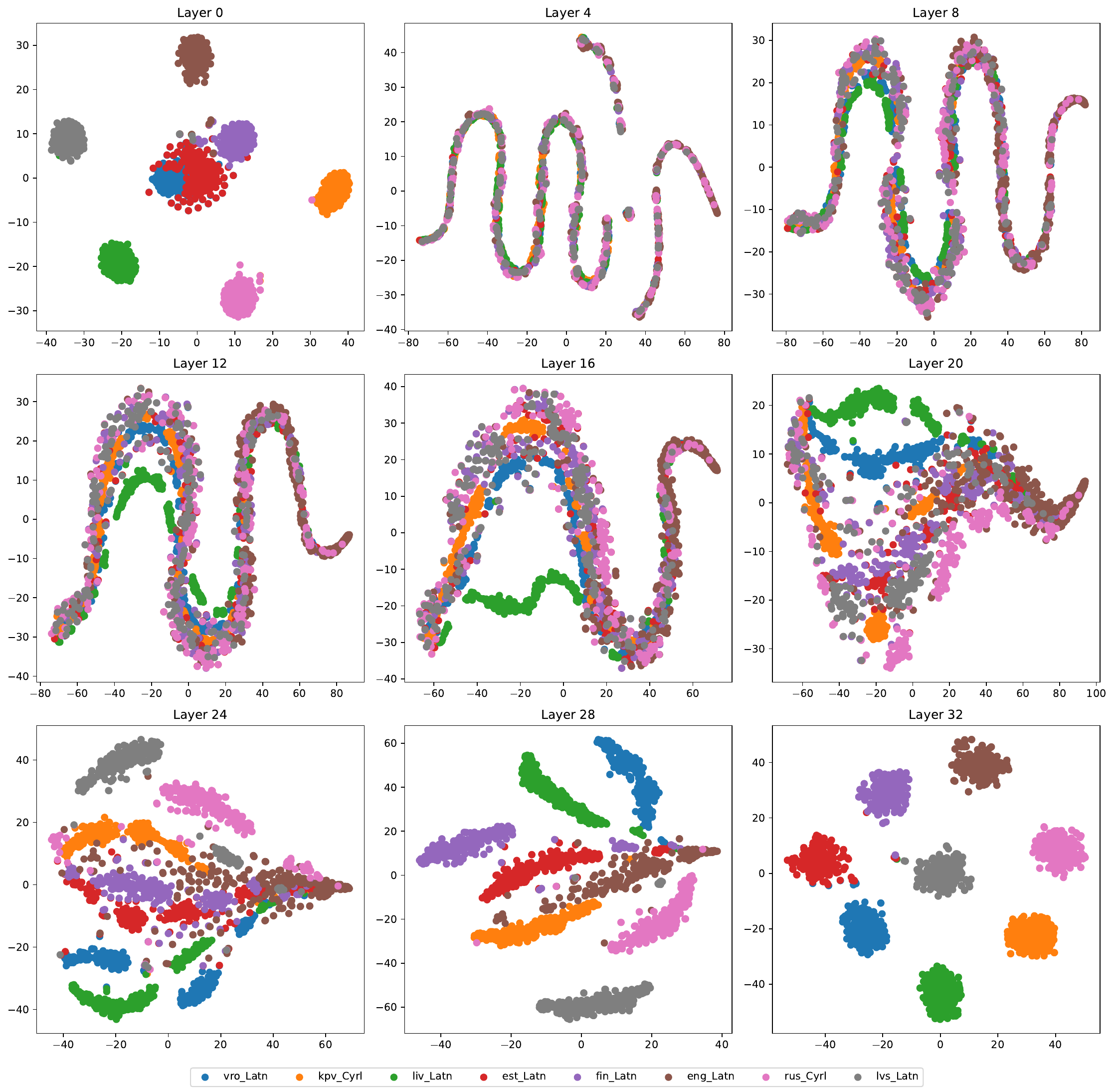}
    \caption{Llama-SMUGRI (Stage 2 + parallel) t-SNE \cite{JMLR:v9:vandermaaten08a} plots of mean-pooled embeddings. Layer 0 is the output of the embedding layer.}
    \label{fig:tsne}
\end{figure*}

The t-SNE \cite{JMLR:v9:vandermaaten08a} plots in Figure~\ref{fig:tsne} show that quite expectedly, the lexically similar languages Võro, Estonian, and Finnish are close or overlapping in the input embeddings, with the other embeddings being language-specific. In the middle layers, the embeddings become more language-agnostic. In the later layers, each language forms a separate cluster. t-SNE plots of Layer 16 embeddings at the end of different training stages suggest representations becoming more language-agnostic as the training progresses. 

\section{Do We Still Need Human Translations?}\label{sec:mt-bench-qe}

We evaluate the applicability of the best proprietary LLMs for creating an MT-Bench-like evaluation dataset for XLR Finno-Ugric languages by asking the models to translate English or Estonian prompts from MT-bench-SMUGRI to the target languages. Table \ref{table:MT-bench-BLEU} shows that the best OpenAI models have not yet learned to translate to XLR Finno-Ugric languages. From that it can be concluded that they could also not generate synthetic data for our target languages with sufficiently good quality.

\begin{table}[!htp]\centering
\scriptsize
\begin{tabular}{lrrrr}\toprule
&\textbf{EST}$\rightarrow$\textbf{VRO} &\textbf{EST}$\rightarrow$\textbf{LIV} &\textbf{ENG}$\rightarrow$\textbf{KPV} \\\midrule
gpt-4o-mini-2024-07-18 &9.3 &5 &4.6 \\
gpt-4o-2024-08-06 &4.5 &5.6 &4.2 \\
gpt-4-turbo-2024-04-09 &18.9 &5.9 &3.6 \\
Neurotõlge &24.7 &21.4 &\textbf{31.7} \\
Llama-SMUGRI-translate &\textbf{26.4} &\textbf{25.3} &19.1 \\
\bottomrule
\end{tabular}
\caption{BLEU scores of translating MT-bench-SMUGRI.}
\label{table:MT-bench-BLEU}
\end{table}

\begin{figure}[h!]
    \centering
        \centering
        \includegraphics[width=0.49\textwidth]{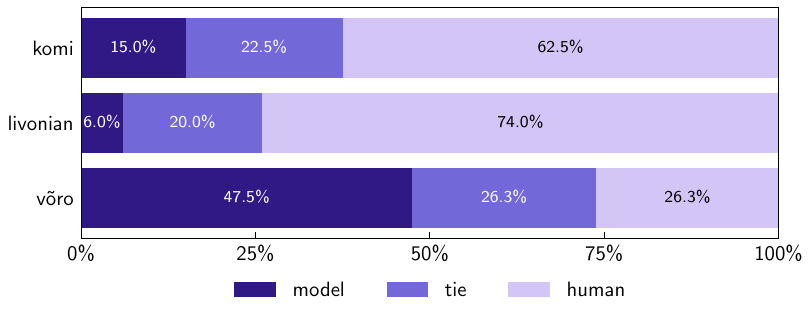}
    \caption{Preferred translations according to pairwise comparison.}
    \label{fig:pairwise-comparison}
\end{figure}

We then compare translations from the best translation models with human translations using pairwise comparison where we ask human annotators to choose a better translation from the two alternatives (ties allowed). We gather 3 sets of annotatios for Livonian, 2 sets for Võro and 1 for Komi. Figure \ref{fig:pairwise-comparison} shows that Komi and Livonian speakers mostly prefer human translations over machine translated data, however, Võro speaker prefer surprisingly often machine translated data suggesting a good quality of Võro machine translation. The average agreement between the pairs of Livonian annotations were 67.5\% while between Võro annotations 42.5\%.

\begin{figure*}
  \centering
   \newtcolorbox{alpacabox}{
  colback=white,
  colframe=black,
  arc=4pt,
  boxrule=0.5pt,
  left=6pt,
  right=6pt,
}
\begin{minipage}{\textwidth}
\begin{alpacabox}
\begin{small}
\textsc{\normalsize PRE-TRAINED MODELS}\\
\rule{\textwidth}{0.1pt} \\
\textbf{FLORES-SMUGRI} \\
\texttt{
\{src\_lang\}: \{src\}\textbackslash n\{tgt\_lang\}:
}\\
\rule{\textwidth}{0.1pt} \\
\textbf{BELEBELE-SMUGRI} \cite[prompt from][]{bandarkar2023belebele} \\
\texttt{
P: \{passage\}\textbackslash nQ: \{question\}\textbackslash nA: \{answer\_1\}\textbackslash nB: \{answer\_2\}\textbackslash nC: \{answer\_3\}\textbackslash nD: \{answer\_4\}\textbackslash nAnswer:}\\
\rule{\textwidth}{0.1pt} \\
\textbf{SIB-SMUGRI} \cite[prompt from][]{lin2024mala500, csaki2024sambalingo} \\
\texttt{
Topic Classification: science/technology, travel, politics, sports, health, entertainment, geography.\textbackslash n\textbackslash nThe label of [\{sentence\}] is}
\\\rule{\textwidth}{0.1pt}\\\\
\textsc{\normalsize INSTRUCTION-TUNED MODELS}\\
\rule{\textwidth}{0.1pt}\\
\textbf{FLORES-SMUGRI}\\
\texttt{
Translate the following \{src\_lang\} text into \{tgt\_lang\}.\textbackslash n\{src\}
} \\
\rule{\textwidth}{0.1pt}\\
\textbf{BELEBELE-SMUGRI} \cite[prompt from][]{bandarkar2023belebele}\\
\texttt{
Given the following passage, query, and answer choices, output the letter corresponding to the correct answer.\textbackslash n\#\#\#\textbackslash nPassage:\textbackslash n\{passage\}\textbackslash n\#\#\#\textbackslash nQuery:\textbackslash n\{query\}\textbackslash n\#\#\#\textbackslash nChoices:\textbackslash n(A) \{answer\_1\}\textbackslash n(B) \{answer\_2\}\textbackslash n(C) \{answer\_3\}\textbackslash n(D) \{answer\_4\}\textbackslash n\#\#\#\textbackslash nAnswer:
} \\
\rule{\textwidth}{0.1pt}\\
\textbf{SIB-SMUGRI} \cite[prompt from][]{adelani-etal-2024-sib} \\
\texttt{
Is this a piece of news regarding science/technology, travel, politics, sports, health, entertainment, or geography?\textbackslash n\{sentence\}
} \\
\rule{\textwidth}{0.1pt}\\
\textbf{GPT-4-turbo as a Fallback Evaluator} (for SIB-200)\\
\texttt{
Your task is to verify if the given model output classifies a text correctly. Answers in other languages should be allowed if they meaning matches closely with the expected class (e.g. \textbackslash See on teadusuudis\textbackslash " is correct when expected output is \textbackslash "science/technology\textbackslash ").  If the model output does not choose a specific class, then the output is incorrect.\textbackslash n\textbackslash n\#\#\# Expected class: \{expected\_answer\}\textbackslash n\textbackslash n\#\#\# Model output: \{output\_text\}\textbackslash n\textbackslash n\#\#\# Respond with Yes or No:
} \\
\end{small}
\end{alpacabox}
\end{minipage}
  \caption{Prompts used for evaluation. Pre-trained models were evaluated with Language Model Evaluation Harness \cite{eval-harness}.}\label{fig:eval-prompts}
\end{figure*}

\begin{figure*}[ht]
    \centering
        \centering
        \includegraphics[width=\textwidth]{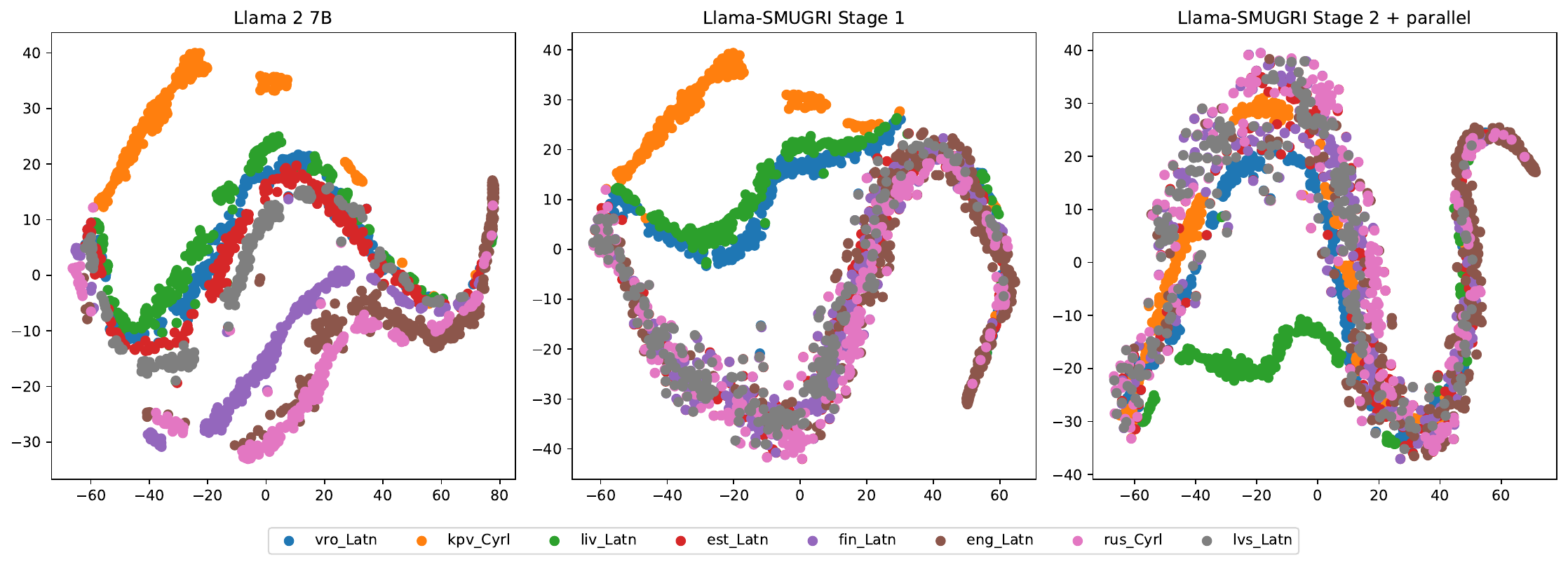}
    \caption{t-SNE \cite{JMLR:v9:vandermaaten08a} plots of mean-pooled 16th layer embeddings in different stages of model development.}
    
\end{figure*}

\end{document}